\documentclass{article}
\usepackage{arxiv}
\usepackage{booktabs}       
\usepackage{amsfonts}       
\usepackage{url,hyperref,lineno,microtype}
\usepackage[]{nomencl}
\usepackage[onehalfspacing]{setspace}
\usepackage[ruled,vlined ]{algorithm2e}
\usepackage{amsmath,amssymb}
\usepackage{mathtools}
\usepackage{float}
\usepackage{graphicx, lscape}
\usepackage{varioref}                 
\usepackage[ruled,vlined]{algorithm2e}
\usepackage{natbib}
\setcitestyle{authoryear,open={(},close={)}}
\title{Deep Reinforcement Learning Controller for 3D Path-following and Collision Avoidance by Autonomous Underwater Vehicles}

\author{
	Simen Theie Havenstrøm  \\
	Department of Engineering Cybernetics\\
	Norwegian University of Science and Technology\\
	Trondheim, Norway \\
	\texttt{simentheie@gmail.com} \\
	\And
	Adil Rasheed
	\thanks{Corresponding author (adil.rasheed@ntnu.no)} \\
	Department of Engineering Cybernetics\\
	Norwegian University of Science and Technology\\
	Trondheim, Norway \\
	\texttt{adil.rasheed@ntnu.no} \\
	\And
	Omer San \\
	School of Mechanical and Aerospace Engineering\\
	Oklahoma State University \\
	Stillwater, Oklahoma, 74078-5016 USA \\
	\texttt{osan@okstate.edu} \\
}

\begin{document}
\maketitle

\begin{abstract}
Control theory provides engineers with a multitude of tools to design controllers that manipulate the closed-loop behavior and stability of dynamical systems. These methods rely heavily on insights about the mathematical model governing the physical system. However, in complex systems, such as autonomous underwater vehicles performing the dual objective of path-following and collision avoidance, decision making becomes non-trivial. We propose a solution using state-of-the-art Deep Reinforcement Learning (DRL) techniques, to develop autonomous agents capable of achieving this hybrid objective without having à priori knowledge about the goal or the environment. Our results demonstrate the viability of DRL in path-following and avoiding collisions toward achieving human-level decision making in autonomous vehicle systems within extreme obstacle configurations.
\end{abstract}
\section{Introduction}\label{sec:introduction}
Autonomous Underwater Vehicles (AUVs) \nomenclature{AUV}{Autonomous Underwater Vehicle} are used in many subsea commercial applications such as seafloor mapping, inspection of pipelines and subsea structures, ocean exploration, environmental monitoring and various research operations. The wide range of operational contexts implies that truly autonomous vehicles must be able to follow spatial trajectories (path following), avoid collisions along these trajectories (collision avoidance \nomenclature{COLAV}{Collision Avoidance}) and maintain a desired velocity profile (velocity control). In addition, AUVs are often underactuated by the fact that they operate with three generalized actuators (propeller, elevation and rudder fins) in six degrees-of-freedom (6-DOF) \nomenclature{6-DOF}{Six Degrees Of Freedom} \citep{fossen2011}. This is the configuration considered in the current work. 

\indent The complexity that arises when combining the control objectives, a complicated hydrodynamic environment and disturbances, and the physical design with three generalized actuators, spurs an intriguing control challenge for which many scientific literature exist. However, the control objectives are in most research dealt with separately.

\subsection{Path Following} \label{ssec:intro_pf}
The path following problem is heavily researched and documented in classical control literature. The control objective is to follow a predefined path, defined relative to some inertial frame, and minimize tracking errors, i.e. the distance between the vehicle and the path. Three-dimensional (3D) \nomenclature{3D}{Three-dimensional} path following involves tracking errors that are composed of horizontal and vertical components, and forms an accurate representation of real engineering operations for AUVs \citep{chu2015}. Typically, a variant of the Proportional Integral Derivative (PID) \nomenclature{PID}{Proportional Integral Derivative} controller based on reduced order models (ROM) is used to control elevator and rudder to eliminate tracking errors \citep[ch.~12]{fossen2011}. More advanced approaches are also available; A classical nonlinear approach is found in \cite{encarnacao2000}, where a kinematic controller was designed based on Lyapunov theory and integrator backstepping. To extend the nonlinear approach reliably in the presence of disturbances and parametric uncertainties, \cite{chu2015} proposed an adaptive sliding mode controller, where an adaptive control law is implemented using a radial basis function neural network. To alleviate chattering, a well-known``zig-zag" phenomenon occurring when implementing sliding mode controllers due to a finite sampling time, an adaptation rate was selected based on a so-called minimum disturbance estimate. \cite{xiang2017} proposed fuzzy logic for adaptive tuning of a feedback linearization PID controller. The heuristic, adaptive scheme accounts for modelling errors and time-varying disturbances. They also compare the performance on 3D path following with conventional PID and non-adaptive backstepping-based controllers, both tuned with inaccurate and accurate model parameters, to demonstrate the robust performance of the suggested controller. \cite{liang2018} suggested using fuzzy backstepping sliding mode control to tackle the control problem. Here, the fuzzy logic was used to approximate terms for the nonlinear uncertainties and disturbances, specifically for use in the update laws for the controller design parameters. Many other methods exist, but most published work on the 3D path following problem incorporate either fuzzy logic, variants of PID control, backstepping techniques or any combination thereof. More recently, there have been numerous attempts to achieve path following and motion control for AUVs by applying machine learning directly to low-level control. Specifically, Deep Reinforcement Learning (DRL) \nomenclature{DRL}{Deep Reinforcement Learning} seems to be the favored approach. DRL controllers are based on experience gained from self-play or exploration, using algorithms that can learn to execute tasks by reinforcing good actions based on a performance metric. In-depth theory on DRL is presented in \autoref{ssec:DRL}.
\cite{yu2017} used a DRL algorithm known as Deep Deterministic Policy Gradients (DDPG) \nomenclature{DDPG}{Deep Deterministic Policy Gradients} \citep{lillicrap2015} to obtain a controller that outperformed PID on trajectory tracking for AUVs. A DRL Controller for underactated marine vessels was implemented in \cite{martinsen2018} to achieve path following for straight-line paths, and later in \cite{martinsen2018_2} for curved paths using transfer learning from the first study. The DRL controller demonstrated excellent performance, even compared to traditional Line-Of-Sight (LOS) \nomenclature{LOS}{Line-Of-Sight} guidance. Exciting results validating the real-world applications of DRL controllers for AUVs and unmanned surface vehicles is found in \cite{carlucho2018} and \cite{woo2019}. The first paper implemented the controller on an AUV equipped with six thrusters configured to generate actuation in pitch moment, yaw moment and surge force. They demonstrated velocity control in both linear and angular velocities. The latter paper implemented a DRL controller on an unmanned surface vehicle with path-following as the control objective, and presented impressive experimental results from the full-scale test.

Common for the aforementioned work published on path-following using DRL controllers are only horizontal motion, i.e. the 2D path following problem, has been considered, and all the works used DDPG as the learning algorithm. The motivation thus lies in utilizing DRL controllers to solve the 3D path following problem is unexplored territory. Moreover, examining how the state-of-the-art DRL algortithm known as Policy Proximal Optimization (PPO)\nomenclature{PPO}{Policy Proximal Optimization}, proposed by \cite{schulman2017}, performs on continuous control problems that has real-world engineering applications is of scientific value. How to setup such training processes and understand how DRL agents learn useful control laws, can provide insights into dynamic systems from a new perspective. 
\subsection{Collision Avoidance}
Collision Avoidance (COLAV) system is an important part of the control systems for all types of autonomous vehicles. AUVs are costly to produce and typically equipped with expensive gears as well. Needless to say, maximum efforts must be made to ensure safe movement at all times. COLAV systems must be able to do \textit{obstacle detection} using sensor data and information processing, and \textit{obstacle avoidance} by applying steering commands based on detection and avoidance logic. The two fundamental perspectives of COLAV control architectures are described in the literature: \textit{deliberate} and \textit{reactive}. \citep{tan2006}. Deliberate architectures are plan driven and therefore necessitates á priori information about the environment and terrain. It could be integrated as part of the on-board guidance system \citep{mcgann2008}, or at an even higher level in the control architecture, such as a waypoint planner \citep{ataei2015}. Popular methods to solve the path planning problem includes A* algorithms \citep{carrol1992, garau2015}, genetic algorithms \citep{Sugihara96} and Probabilistic roadmaps\citep{kavraki1996, cashmore2014}. Deliberate methods are computationally expensive, due to information processing about the global environment. However, they are more likely to make the vehicle converge to the objective \citep{eriksen2016}. Reactive methods on the other hand, are faster and processes only real-time sensor data to make decisions. In this sense, the reactive methods are considered local and are used when rapid action is required. Examples of reactive methods are the dynamic window approach \citep{fox1997, eriksen2016}, artificial potential fields \citep{williams1990} and constant avoidance angle \citep{wiig2018}. A potential pit-fall with reactive methods are local minima manifested as dead-ends \citep{eriksen2016}. To improve on both the deliberate and the reactive approach, a \textit{hybrid} approach is used in practice by combining the strengths of both. Such architectures are comprised of a deliberate, reactive and execution layer. The deliberate layer handles high level planning, while the reactive layer tackles incidents happening in real-time. The execution layer facilitates the interaction between the deliberate and reactive architectures and decides the final commanded steering. \citep{tan2006} 

There are still challenges in state-of-the-art COLAV methods for vehicles subjected to nonholonomic constraints, such as AUVs. Instability issues, neglecting vehicle dynamics and actuator constraints leading to infeasible reference paths, and algorithms causing the vehicle to stop are recurring challenges seen in the literature. Additionally, extensive research discusses methods for COLAV in 2D that cannot be directly applied to 3D. In many cases where such methods are adapted to 3D, however, they do not optimally take advantage of the extra dimension \citep{wiig2018}.

\subsection{Research Goals and Methods}\label{ssec:Objective}
In this research, we attempt to achieve the control objectives by employing a DRL controller as the motion control system operating the control fins. The level of complexity of the hybrid control problem suggests using an intelligent controller, such as a DRL agent, to learn a control law through exploration. The agent commands the control fins, while a traditional PI-controller maintains a desired cruise speed. The key idea lies in the fact that the agent learns operating both the elevator and rudder at the same time, and should therefore be able to learn an optimal strategy for navigating in both planes. The challenge of DRL control is establishing a reward function such that safe and effective tracking behaviour is incentivized.

In addition to setting up a DRL environment where learning happens through exploration and feedback through a reward signal, the learning strategy known as \textit{curriculum learning} is employed: That is the formalization of learning by being gradually and systematically exposed to more complex environments \citep{bengio2009}. As the control objectives can be well-described in terms of complexity, specifically the density of obstacles blocking the path or the intensity of an external disturbance, it is a viable approach in this context. Note that any arbitrary scenario configuring the path and obstacles can be generated, so another key component in the research is designing meaningful configurations in a practical sense. If this is achieved, any agent that has been training in simulation could in theory be uploaded to a physical unit and continue learning in a full-scale test environment. 

To implement curriculum learning, scenarios ranging from \textit{beginner} to \textit{expert} level difficulty are constructed. Initially, we start with only a path and no obstacles or ocean current disturbance and train until the agent master that difficulty. Then, obstacles are progressively added and eventually an ocean current disturbance is introduced to form the expert level scenario. Scenarios are detailed in \autoref{ssec:scenarios}. In a COLAV sense, the predefined path can be viewed as the deliberate architecture, where it is assumed that the waypoints are generated by some path planning scheme, and the random and unforeseen obstacles are placed on this presumed collision-free path. The DRL agent operates in effect as the reactive system that must handle the threat of collisions rapidly, but at the same time chooses effective trajectories to reach the target. 

\section{Theory}\label{sec:theory}
\subsection{AUV Model}\label{ssec:auv_model}
The equations of motion for the AUV are sepearted into kinematics and kinetics, represented by the state vectors $\boldsymbol{\eta} = [\boldsymbol{\mathbf{p}^n}, \boldsymbol{\Theta}_{nb}]^T =  [x,y,z,\phi,\theta,\psi]^T$ and $\boldsymbol{\nu} = [\mathbf{v}^b,\boldsymbol{\omega}_{b/n}^b]^T = [u,v,w,p,q,r]^T$, respectively. The components are defined in \autoref{tab:notation} in accordance with the notation given by the Society of Naval Architects and Marine Engineers (SNAME (1950)). We consider the dynamics in a vehicle fixed BODY-frame (written $\{b\}$), that is the reference frame centered at the AUV's center of control, and the North-East-Down frame (written $\{n\}$), which is considered inertial so that Newton's laws of motion applies. Transformations between the two reference frames are given by the the rotation matrix $\mathbf{R}^n_b(\boldsymbol{\Theta}_{nb})$ following the z-y-x convention. \citep{fossen2011}

\begin{table*}[h]
\centering 
\caption{Notation for marine vessels as given by SNAME (1950)}
\begin{tabular}{||l|c|c|c|c|| }
\hline
\textbf{Degree of freedom} & \textbf{Forces and moments} & \textbf{Velocities} & \textbf{Positions} \\
\hline
1 translation in the $x$ direction (surge)  & $X$ & $u$ & $x$ \\
2 translation in the $y$ direction (sway)   & $Y$ & $v$ & $y$ \\
3 translation in the $z$ direction (heave)  & $Z$ & $w$ & $z$ \\
4 rotation about $x$ axis (roll)        & $K$ & $q$ & $\phi$ \\
5 rotation about $y$ axis (pitch)       & $M$ & $p$ & $\theta$ \\
6 rotation about $z$ axis (yaw)         & $N$ & $r$ & $\psi$ \\
\hline
\end{tabular}
\label{tab:notation}
\end{table*}

The kinematic state vector is the position of the vessel concatenated with the attitude w.r.t. $\{n\}$. Since the vessel's velocity is defined in $\{b\}$, a differential equation for $\boldsymbol{\eta}$ is obtained by transforming $\boldsymbol{\nu}$ as seen in \autoref{eq:kinematicequations}.

\begin{equation}
    \Dot{\boldsymbol{\eta}} = 
    \begin{bmatrix}
    \Dot{\mathbf{p}}^n \\
    \Dot{\boldsymbol{\Theta}}_{nb}
    \end{bmatrix} =
    \begin{bmatrix}
    \mathbf{R}_b^n(\boldsymbol{\Theta}_{nb}) & \mathbf{0} \\
    \mathbf{0} & \mathbf{T}_{\Theta}(\boldsymbol{\Theta}_{nb})
    \end{bmatrix} 
    \begin{bmatrix}
    \mathbf{v}^b \\
    \boldsymbol{\omega}_{b/n}^b    
    \end{bmatrix}=
    \mathbf{J}_{\Theta}(\boldsymbol{\eta}) \boldsymbol{\nu}
    \label{eq:kinematicequations}
\end{equation}

Here, the transformation $\mathbf{T}_\Theta(\boldsymbol{\Theta}_{nb})$ relating the angular velocities to the Euler angles is given by \autoref{eq:transformation}.

\begin{equation}
    \mathbf{T}_\Theta(\boldsymbol{\Theta}_{nb})
    = 
    \begin{bmatrix}
    1 & s\phi t\theta & c\phi t\theta \\
    0 & c\phi & -s\phi \\
    0 & \frac{s\phi}{c\theta} & \frac{c\phi}{c\theta}
    \end{bmatrix}
    \label{eq:transformation}
\end{equation}

Some key assumptions about the AUV must be made in order for the kinetic equation to be valid. We assume that 1) the AUV operates at a depth where disturbances from wind and waves are negligible; 2) the maximum speed is $2m/s$; 3) mass is distributed such that the moments of inertia can be approximated by that of a spheroid; 4) the center of gravity is located $1cm$ under the center of buoyancy to create a restoring moment in roll and pitch; 5) the AUV is top-bottom and port-starboard symmetric; 6) the AUV is slightly buoyant, as a fail-safe mode in case of power loss. The model and its parameters are adapted from \cite{dasilva2007}, where a detailed walk-through of the kinetic equation matrices, the parameters and AUV specifications used in the simulation model is offered. 

To simulate the kinetic equations, a 6-DOF nonlinear model is implemented. Hydrodynamics are notoriously difficult to represent accurately, so that heuristical formulas are often reverted to. Doing this implies that the AUV kinetics can be described as a mass-spring-damper system, according to \autoref{eq:kineticequations}: 

\begin{equation}\label{eq:kineticequations}
    \underbrace{\mathbf{M}\Dot{\boldsymbol{\nu}}}_{Mass forces} +
    \underbrace{\mathbf{C}(\boldsymbol{\nu})\boldsymbol{\nu}}_{Coriolis forces} + 
    \underbrace{\mathbf{D}(\boldsymbol{\nu})\boldsymbol{\nu}}_{Damping forces} +
    \underbrace{\mathbf{g(\boldsymbol{\eta}})}_{Restoring forces} = 
    \boldsymbol{\tau}_{control}
\end{equation}

The definitions for these matrices are standard, but note that lift is included in the term $\mathbf{D}(\boldsymbol{\nu})$. Moreover, the damping component contains linear and quadratic terms to emulate linear viscous damping and nonlinear damping due to phenomena such as vortex shedding. The control force vector $\boldsymbol{\tau}_{control}$ is a function of the three actuators propeller shaft speed, rudder (vertical fin) angle and elevator (horizontal fin) angle, written $\eta, \delta_r, \delta_s$, respectively. These actuators are all constrained, so that $0 \leq \eta \leq \eta_{max}$ and $|\delta_r|, |\delta_s| \leq \delta_{max}$. To not violate the low-speed assumption, $\eta_{max}$ must be chosen accordingly. 

To simulate operation of the control fins more realistically, the output of the DRL controller passes through a first-order low-pass filter with time-constant $T_f$. The intention behind this implementation is to remove noisy outputs from the DRL agent, without having to add a cost to the control action derivatives $\dot{\delta}_r$ and $\dot{\delta}_s$. Ideally, the agent learns that abrupt control action is pointless since high frequency commands are filtered out. The implementation of the discrete low-pass filter for the fins is given by \autoref{eq:lowpass}:

\begin{equation}
    \delta_{r, t} = (1-a) \delta_{r, t-1} + a u_t 
    \label{eq:lowpass}
\end{equation}

\noindent where the filter-parameter $a$ is related to the time-constant by $a = \frac{h}{T_f+h}$, $u_t$ is the raw control action and $h$ is the step-size. \citep{haugen2008}

Lastly, we present the simulation model used for the ocean current disturbance. This is based on generating the intensity of the current, $V_c = \Vert \boldsymbol{\nu}_c \Vert_2 $, by utilizing a first-order \textit{Gauss-Markov Process} \citep{fossen2011}: 

\begin{equation}
    \dot{V}_c = -\mu V_c + w    
\end{equation}

\noindent where $w$ is \textit{white noise} and $\mu \geq 0$ a constant. An integration limit is set so that the current speed is limited between $0.5$ to $1~m/s$. The current direction is static and initialized randomly for each episode. The current direction is described by the sideslip angle and angle of attack, symbolized by $\alpha_c$ and $\beta_c$, respectively, representing at what direction the current hits the body frame. In NED coordinates, the linear ocean current velocities can be obtained by \autoref{eq:currentned}. \citep{fossen2011}

\begin{equation}\label{eq:currentned}
    \mathbf{v}^n_c = V_c \begin{bmatrix}
    \cos{\alpha_c} \cos{\beta_c} \\
    \sin{\beta_c} \\
    \sin{\alpha_c} \cos{\beta_c}
    \end{bmatrix}
\end{equation}

There are no dynamics associated with the sideslip angle and the angle of attack in the simulations; the current direction stays fixed throughout an episode. To obtain the linear velocities in the body frame, we apply the inverse Euler-angle rotation matrix, as seen in \autoref{eq:currentbody}:

\begin{equation}\label{eq:currentbody}
    \begin{bmatrix} 
    u_c \\
    v_c \\
    w_c
    \end{bmatrix} = 
    \mathbf{R}_b^n(\boldsymbol{\Theta}_{nb})^T \mathbf{v_c^n}
\end{equation}

Since the ocean current is defined to be irrotational, the full current velocity vector is written $\boldsymbol{\nu}_c = [u_c, v_c, w_c, 0, 0, 0]$. The current is included by simply subtracting the current velocity from the AUV velocity such that $\boldsymbol{\nu}_r = \boldsymbol{\nu} - \boldsymbol{\nu}_c$ and simulate the dynamics for $\boldsymbol{\nu}_r$ according to \autoref{eq:kineticequations}. This can be done since we use a parametrization of the coriolis matrix that is independent of linear velocites and the ocean current is irrotational \citep{fossen2011}. 

\subsection{Deep Reinforcement Learning}\label{ssec:DRL}
In RL an algorithm, known as an \textbf{agent}, makes an \textbf{observation} $s_t$ of an \textbf{environment} and performs an \textbf{action} $a_t$. The observation is referred to as the state of the system, and is drawn from the state space $\mathcal{S}$. The action is restricted to the well-defined action space $\mathcal{A}$. When an RL task is not infinitely long, but ends at some time $T$, we say that the problem is episodic, and that each iteration through the task is an episode.

After performing an action the agent receives a scalar \textbf{reward} signal $r_t=r(s_t,a_t)$. The reward quantifies how good it was to choose action $a_t$ when in state $s_t$. The objective of the agent is typically to maximize expected cumulative reward. The action choices of the agent are guided by a \textbf{policy} $\pi(s)$, which can be either deterministic or stochastic. In the case that the learning algorithm involves a neural network, the policy is parametrized by the learnable parameters of the network, denoted by $\boldsymbol{\theta}$. When the policy is stochastic and dependent on a neural network, we write $\pi(s) = \pi_{\boldsymbol{\theta}}(a|s)$.

\subsubsection{Proximal Policy Optimization}
The actor-critic algorithm known as Proximal Policy Optimization was proposed by \cite{schulman2017}. We briefly present the general theory and the algorithm which is used in this research. Let the value-function $V(s)$ represent the expected cumulative reward during and episode when following the current policy. In addition, let the state-action value-function $Q(a,s)$ define the expected cumulative reward by following the policy and by taking initial action $a$. Then the advantage function $A(s,a)$ is given by:
\begin{equation}
    A(s,a) = Q(s,a) - V(s).
\end{equation}
\noindent The advantage function represents the difference in expected return by taking action $a$ in state $s$, as opposed to following the policy. Because both $Q(s,a)$ and $V(s)$ are unknown, an estimate of the advantage function, $\hat{A_t}$, is calculated based on an estimate of the value function $\hat{V}(s)$, which is made by the critic neural network. When the value-function is estimated, an alternative for estimating the advantage function is the generalized advantage estimate (GAE), given in \autoref{eq:gae}~\cite{schulman2015}.
\begin{equation}\label{eq:gae}
\begin{gathered}
    \hat{A}_t = \delta_t + (\gamma \lambda) \delta_{t+1} + \dots + (\gamma \lambda)^{T-t+1} \delta_{T-1} \\
    \text{where } \delta_t = r_t + \gamma \hat{V}(s_{t+1}) - \hat{V}(s_t)
\end{gathered}
\end{equation}

Here, $T$ is a truncation point which is typically much smaller than the duration of an entire episode. As before, $\gamma$ is the discount factor. As the GAE is a sum of uncertain terms, the tuneable parameter $0 \leq \lambda \leq 1$ is introduced to reduce variance. However, $\lambda < 1$ makes the GAE biased towards the earlier estimates of the advantage function. Hence, choosing $\lambda$ is a bias-variance trade-off.

The second key component in PPO is introducing a surrogate objective. It is hard to apply gradient ascent directly to the RL objective. Therefore, Schulman et al. suggest a surrogate objective which is such that an increase in the surrogate provably leads to an increase in the original objective ~\cite{schulman2017}. The proposed surrogate objective function is given by \autoref{eq:PPOobjective}.
\begin{equation}\label{eq:PPOobjective}
\resizebox{0.8\linewidth}{!}{$
    L^{CLIP}(\theta) = \hat{\mathbb{E}}_t [ \min \left( \frac{\pi_{\theta}(a_t | s_t)}{\pi_{\theta_{old}}(a_t | s_t)}\hat{A}_t,  clip \left( \frac{\pi_{\theta}(a_t | s_t)}{\pi_{\theta_{old}}(a_t | s_t)},1-\epsilon, 1+\epsilon \right) \hat{A}_t \right) ]  $}
\end{equation}

The tuning parameter $\epsilon$ reduces the incentive to make very large changes to the policy at every step of the gradient ascent. This is necessary as the surrogate objective only estimates the original objective locally in a so-called trust-region. During a training iteration, $N$ actors (Parallelized agents) are enabled to execute the policy and in that way sample trajectories for $T$ timesteps. Then the GAE is computed based on the sampled trajectories, and the advantage estimation is used to optimize the surrogate objective for K epochs using mini-batches of size M per update. The PPO method is seen in its general form in \autoref{alg:PPO} \citep{schulman2017}.

\begin{algorithm}[H]
\caption{Proximal Policy Optimization, Actor-Critic style}
\SetAlgoLined
\For{iteration: 1,2...}{
\For{actor: 1,2...N}{
\text{Run policy $\pi_{\boldsymbol{\theta}_{old}}$ for T time-steps} \\
\text{Compute advantage estimate $\hat{A}_1...\hat{A}_T$}
}
\text{Optimize surrogate L w.r.t. $\boldsymbol{\theta}$, with K epochs and mini-batch size  $M<NT$} \\
\textbf{$\boldsymbol{\theta}_{old} \leftarrow \boldsymbol{\theta}$}
}
\label{alg:PPO}
\end{algorithm}

\subsection{Guidance Laws for 3D Path Following} \label{ssec:theory_pf}
In order to avoid the continuity weaknesses of constructing linear path segments, it is important to generate a smooth reference path. There are different ways to achieve this, but we use a 3D extension of Qudratic Polynomial Interpolation (QPMI)\nomenclature{QPMI}{Qudratic Polynomial Interpolation} proposed by \cite{chu2015}. In short, generating a QPMI path in 3D is done by calculating the quadratic polynomial coefficients that links three-and-three waypoints togheter. The number of waypoints $n_w$ and their configuration is arbitrary. Using the QPMI method, one obtains $2-n_w$ polynomials that are curvature continuous in connecting any three waypoints. Naturally, the goal is to link all waypoints together, which is achieved by using a membership function to merge the obtained polynomial functions. The details are not critical for this work, so for an extensive explanation see the original paper by \cite{chu2015}. They also go on to prove curvature continuity in the resulting paths. For a visual example of how the resulting path looks, see \autoref{ssec:scenarios}. 

To define the tracking errors, the Serret-Frenet (\{SF\}) reference frame associated with each point of the path is introduced. The $x_{SF}$ axis points tangent to the path, the $y_{SF}$ axis normal to the path and the $z_{SF}$ axis points orthogonal to both such that $z_{SF} = x_{SF} \times y_{SF}$ \citep{encarnacao2000}. The tracking-error vector, $\boldsymbol{\varepsilon} = [\Bar{s}, e, h]^T$ is defined by the along-track, cross-track and vertical-track error. The tracking-error vector points towards the closest point on the path from the vessel. Because the origin of the \{SF\} frame can be arbitrarily placed, the the point on the path closest to vessel is chosen as the origin in the simulation. This yields $\Bar{s}=0$, which intuitively makes sense in a path following scenario since the path is not dependent on time. There is therefore no error in the along-track distance component. 

\begin{equation}\label{eq:trackingerror}
    \boldsymbol{\varepsilon} = \mathbf{R}^{SF}_n(\upsilon_p, \chi_p)^T(\mathbf{P}^n-\mathbf{P}_p^n)
\end{equation}

\noindent where $\mathbf{P}^n$ is the position of the vessel and $\mathbf{P}^n_p$ is the closest point on the path. Now the desired azimuth and elevation angle can be calculated according to:

\begin{equation}
    \chi_d(e) = \chi_p + \chi_r(e) \;\;\;\;\;,\;\;\;\;\; \upsilon_{d}(h) = \upsilon_p + \upsilon_r(h)
    \label{eq:des_ang}
\end{equation}

\noindent where $\chi_r(e) = \arctan(-\frac{e}{\Delta})$ and $\upsilon_r(h) = \arctan(\frac{h}{\sqrt{e^2 + \Delta^2}})$. It is seen that driving $e$ and $h$ to zero will in turn drive the correction angles $\chi_r(e)$ and $\upsilon_r(h)$ to zero, and the velocity vector then aligns with the tangent of the path given when $\chi = \chi_d = \chi_p$ and $\upsilon = \upsilon_d = \upsilon_p$.

\section{Method and Implementation}\label{Implementation}
The simulation environments are built to comply with the \textbf{OpenAI Gym} \citep{brockman2016} standard interface. For the RL algorithms, the \textbf{Stable Baselines} package which offers improved parallelizable implementations based on \textbf{OpenAI Baselines}\citep{baselines2017} library is utilized. Ten different scenarios are created: Five for testing and five for training. 

\subsection{Environment Scenarios}\label{ssec:scenarios}
Training scenarios are constructed by generating a path from a random set of $n_w$ waypoints which are generated such that unrealistically sharp turns are avoided. The first scenario used in curriculum learning is called \textit{beginner}, where only a path and no obstacles or ocean current is present. Then the agent is introduced to the \textit{intermediate} level, where a single obstacle are placed on the half-way mark. Next level is called \textit{proficient}: Here two more obstacles are placed equally distanced from the half-way mark. 

\begin{figure}[h]
    \centering
    \includegraphics[width=\linewidth]{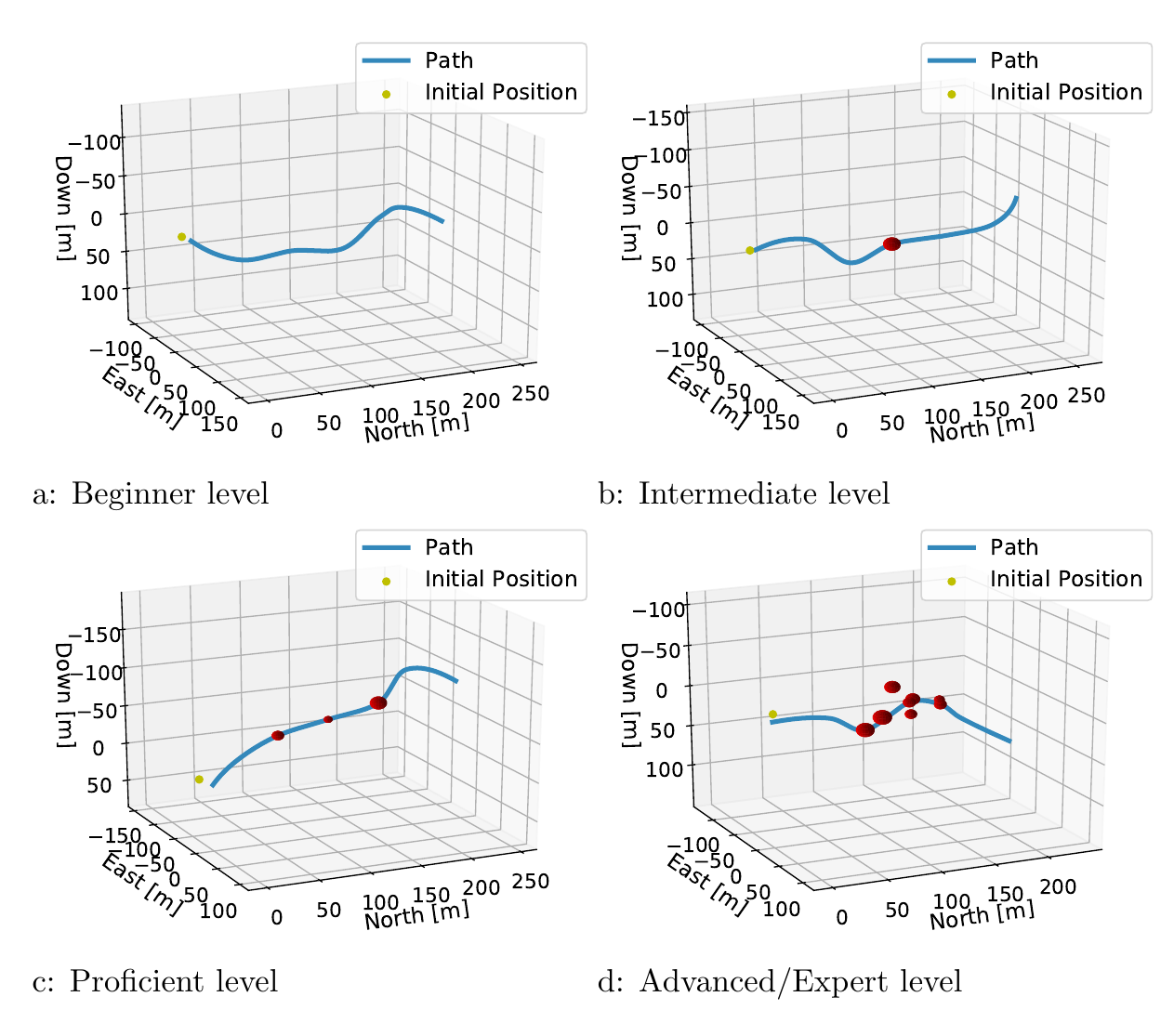}
	\caption{Training Scenarios used in curriculum learning.}
	\label{fig:training_scenarios}
\end{figure}

The last part of training happens in the \textit{advanced} and \textit{expert} level scenarios. In the advanced level difficulty, we generate the \textit{proficient} challenge, but additionally five more obstacles are placed randomly off-path, such that an avoidance maneuver could induce a new collision course. The distinction between the expert and the advanced level is the inclusion of the ocean current disturbance. In all scenarios the first and the last third of the path is collision-free, in order to keep part of curriculum from the beginner scenario (pure path-following) present throughout the learning process. This enables the agent to not forget knowledge learned from doing path-following only.  \autoref{fig:training_scenarios} illustrates the different training contexts the agent is exposed to. 

In addition to train the agent progressively through these scenarios, quantitative evaluation is performed by sampling a number of episodes such that the agents' performance across the various difficulty levels can be established. 

After evaluating the controllers by statistical averages, qualitative analysis is done in designated test scenarios. These are designed to test specific aspects of the agents' behaviour. The first scenario tests pure path-following on a non-random path (in order for results to be reproducible) both with and without the presence of an ocean current. Next, special (extreme) cases where it would be preferable to use only one actuator for COLAV, i.e. horizontally and vertically stacked obstacles, are generated. The agents are also tested in a typical pitfall scenario for reactive COLAV algorithms: A dead-end. See \autoref{ssec:qual_result} for illustrations of the test scenarios. 

\subsection{Obstacle Detection}
\begin{figure}[H]
    \centering
    \includegraphics[width=\linewidth]{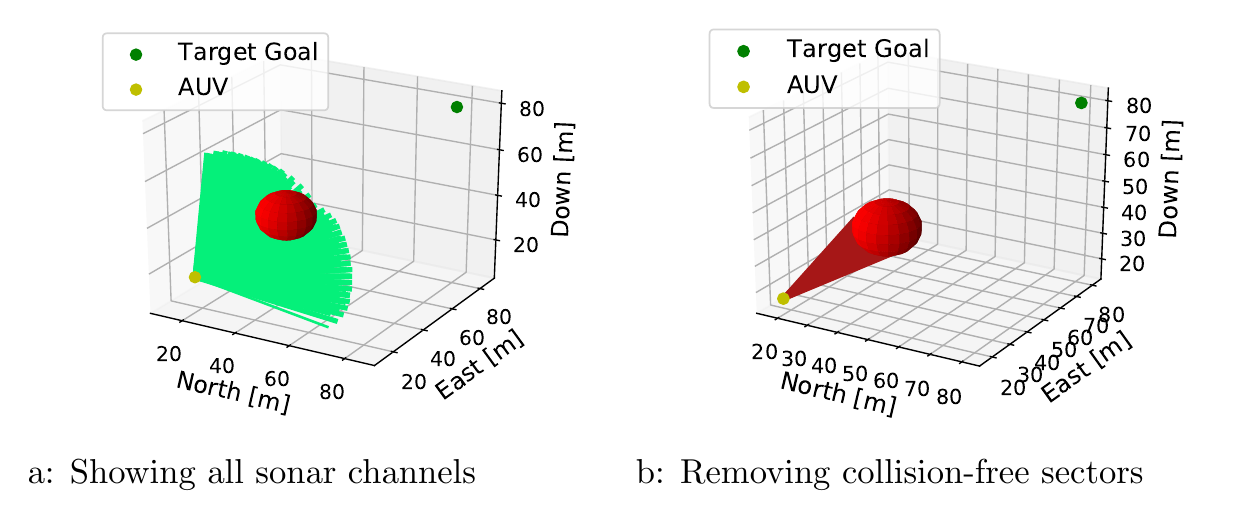}
\caption{Rendering of the sonar simulation during an active episode.}
\label{fig:sonars}
\end{figure}

Being able to react to the unforeseen obstacles require the AUV to perceive the environment through sensory inputs. This perception, or obstacle detection, is simulated by providing the agent a 2D sonar image, representing distance measurements to a potential intersecting object in front of the AUV. This setup emulates a Forward Looking Sonar (FLS)\nomenclature{FLS}{Forward Looking Sonar} mounted on the front of the AUV. A 3D rendering of the FLS simulation is seen in \autoref{fig:sonars}. The specific sensor suite, the sonar range and the sonar apex angle is configurable, and can therefore be thought of as hyperparameters.

Depending on the sensor suite of choice, the number of sensor rays can get quite large. It is also notable that this issue is exponentially larger in 3D compared to 2D, slowing the simulation speed significantly as searching through the sonar rays (line search) is computationally expensive. For this reason, the sensor suite used in this research is $15$ by $15$, providing a grid with $10^{\circ}$ spacing between each sonar ray when scanning with a $140^{\circ}$ apex angle. This amounts to a total of $225$ line searches per sensor update and in order to limit this stress on computational resources, the update frequency is set to $1Hz$. Moreover, the sonar range is limited to $25m$.
\subsection{Reward Function}
Reward function design is a crucial part of any RL process. The goal is to establish an incentive so the agent learns certain behavioural aspects. This is done by trying to imitate human-like behaviour. For instance, following the path is objectively desirable, but this goal must be suspended in the case off a potential collision. When to react and by what safety margin is inherently a subjective choice. Regulating this trade-off is a balancing act, where following the path notoriously would result in many collisions and being too cautious would be ineffective. Additionally, a configuration involving excessive roll, i.e. the angular displacement of the AUV arounds its own longitudinal axis, is undesirable because that implies inverting or even swapping the two actuators' effect (the rudder would operate as the elevator and vise versa) in terms of combating course and elevation errors. Not using the actuators to aggressively is therefore key in achieving smooth and safe operation. Thus, a reward function incorporating these important aspects of AUV motion control is developed. 

The first part focuses on path-following and simply penalizes errors between desired and actual course and elevation angle, as given by \autoref{eq:reward_pf}:

\begin{equation}
    r_t^{pf}(\tilde{\chi}, \tilde{\upsilon}) = c_\chi \tilde{\chi}^2 + c_\upsilon \tilde{\upsilon}^2 
    \label{eq:reward_pf}
\end{equation}

Where $c_\chi$ and $c_\upsilon$ are negative weights deciding the severity of being off the course and elevation angles calculated by the guidance laws. The next incentive is avoiding obstacles blocking the path seen through the 2D sonar image. First, the range measurements are converted to a proportionally inverse quantity we have called \textit{obstacle closeness}. This quantity is written $c(d_{i,j}) = \text{clip} \left (1-\frac{d_{i,j}}{d_{max}}, 0, 1 \right)$, where $d_{i,j}$ is the i'th and j'th pixel distance measurement and $d_{max}$ is the sonar range. This transformation sets all sensor inputs zero as long as there are no obstacles nearby, effectively deactivating learning in this part of the neural net during the beginner scenario. The term incentivizing obstacle avoidance is written in \autoref{eq:reward_colav}. It is calculated as a weighted average in order to remove the dependency on a specific sensor suite configuration. Furthermore, a small constant $\epsilon_c$ is used to remove singularities occurring when obstacle closeness in a sector is exactly $1$ and $\gamma_c$ is a scaling parameter.  

\begin{equation}
    r^{oa}_t(\mathbf{d}) = -\frac{\sum_{i\in\mathcal{I}} \sum_{j\in\mathcal{J}} \beta_{oa}(\theta_j, \psi_i) \left(\gamma_c \max \left((1-c(d_{i,j}))^2, \epsilon_c \right)\right)^{-1}}{\sum_{i\in\mathcal{I}} \sum_{j\in\mathcal{J}} \beta_{oa}(\theta_j, \psi_i)}
    \label{eq:reward_colav}
\end{equation}

\begin{figure}[H]
    \centering
    \includegraphics[width=0.75\linewidth]{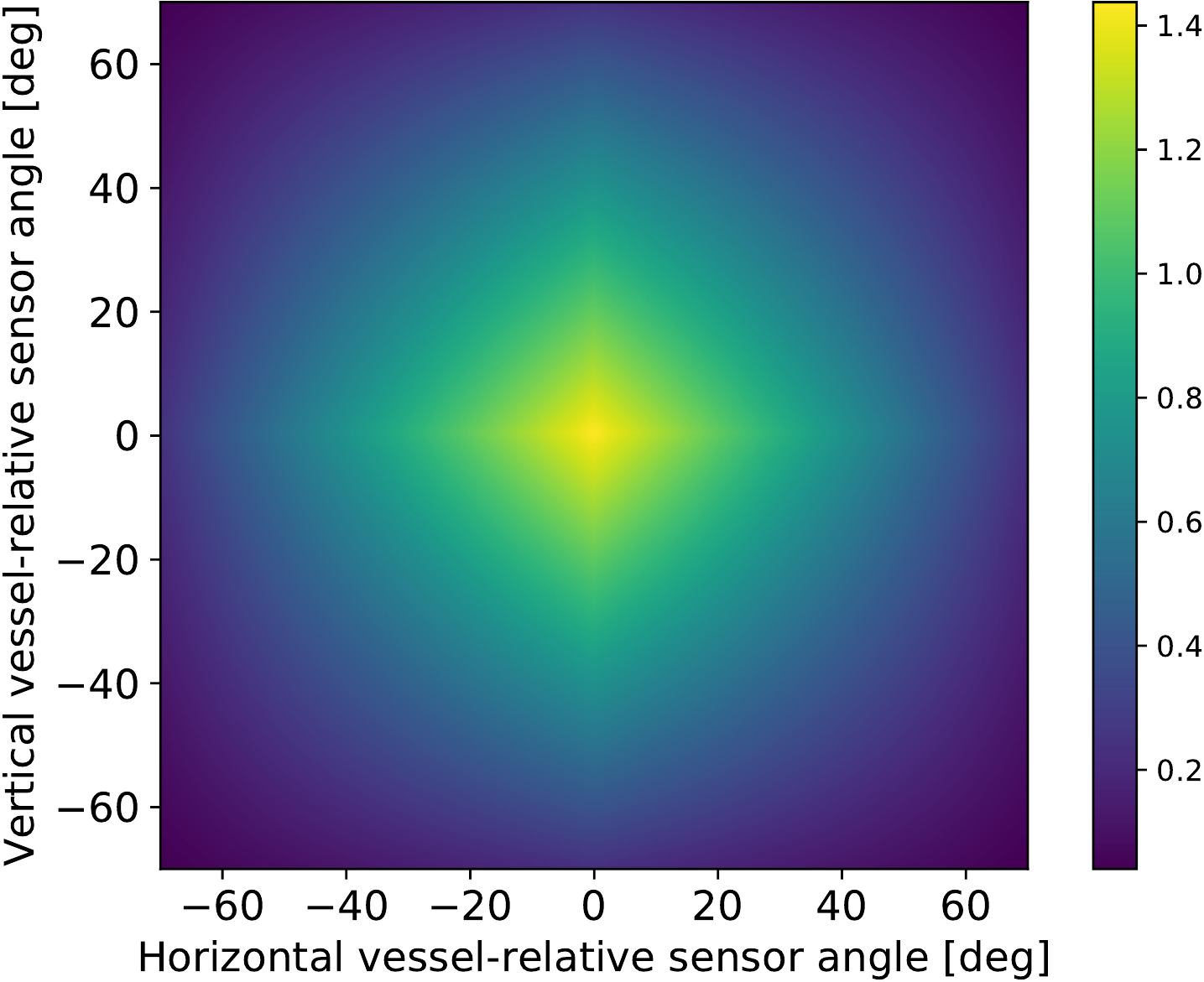}
    \caption[Two-variable function for obstacle reward scaling]{How the reward is scaled according to the sonar-data's vessel-relative direction. Note that the grid illustrated is much finer than the $15$ by $15$ sensor suite used during simulation.}
    \label{fig:obstacle_reward_scalar}
\end{figure}

Since the vessel-relative orientation of an obstacle determines whether a collision is likely, the penalty related to a specific closeness measurement is scaled by an orientation factor dependent on the relative orientation. The vessel-relative scaling factor is written $\beta_{oa}(\theta_j, \psi_i) = (1 - \frac{2|\theta_i|}{\gamma_a})(1 - \frac{2|\psi_j|}{\gamma_a}) + \epsilon_{oa}$. Here, $\epsilon_{oa}$ is a small design constant used to penalize obstacles at the edge of the configuration, and $\theta_j$ and $\psi_j$ defines the vessel-relative sonar direction. \autoref{fig:obstacle_reward_scalar} illustrates how the 2D sonar image is weighted in terms of the sector importance given by $\beta_{oa}$. As is clear, obstacles that appear centermost in the sonar image will yield the largest penalty.

To find the right balance between penalizing being off-track and avoiding obstacles - which are competing objectives - the weight parameter $\lambda_r \in [0,1]$ is used to regulate the trade-off. This structure is adapted from the work by \cite{meyer2019}, which performed the analog experiments in 2D. In addition, we add penalties to roll, roll rate and the use of control actuation to form the complete reward function: 
\begin{equation}
    r_t(\tilde{\chi}, \tilde{\upsilon}, \mathbf{d}, \phi, r, \delta_r, \delta_s) = \lambda_r r^{pf}_t(\tilde{\chi}, \tilde{\upsilon}) + (1-\lambda_r) r^{oa}_t(\mathbf{d}) + c_\phi \phi^2 + c_r r^2 + c_{\delta_r} \delta_r^2 + c_{\delta_s} \delta_s^2
\end{equation}
\subsection{Feedback/Observations}
The list of state observations, referring to the states of the dynamical model, the agents inputs during training and in operation is seen in \autoref{tab:pf_e2e_obs}. The inputs are normalized by the true or the empirical maximum, so that values passed into the neural network is in the range $[-1, 1]$. Input normalization is used to improve the speed of convergence and the symbols are denoted by subscript $o$ to indicate that these are the actual values passed as observations. The nonlinear activation functions of neural networks tend to saturate if the inputs gets too large, hence normalization is a means used to counteract this effect. Furthermore, large input values might lead to huge error gradients, which in turn causes unstable training. Normalization is therefore a simple form of pre-processing contributing to faster and more stable training. \citep{lecun1998}

\vspace{-1cm}

\begingroup
\setlength{\tabcolsep}{6pt} 
\renewcommand{\arraystretch}{1.5} 
\begin{table}[H]
\caption[Observations for end-to-end training]{Observation table for end-to-end training for path following. All states and errors are normalized by the empirical or true maximum value.}
\centering
\begin{tabular}{l c c c}
\hline
\textbf{Observation} & & \textbf{Max} \\
\hline
Relative surge speed & $u_{ro} = \frac{u_r}{u_{max}} \in [-1,1]$ & $2$ \\
Relative sway speed & $v_{ro} = \frac{v_r}{v_{max}} \in [-1,1]$ & $0.3$ \\
Relative heave speed & $w_{ro} = \frac{w_r}{w_{max}} \in [-1,1]$ & $0.3$ \\
Roll & $\phi_o = \frac{\phi}{\phi_{max}} \in [-1,1]$ & $\pi$ \\
Pitch & $\theta_o = \frac{\theta}{\theta_{max}} \in [-1,1]$ & $\pi$ \\
Yaw & $\psi_o = \frac{\psi}{\psi_{max}} \in [-1,1]$ & $\pi$ \\
Roll rate & $p_o =  \frac{p}{p_{max}} \in [-1,1]$ & 1.2 \\
Pitch rate & $q_o = \frac{q}{q_{max}} \in [-1,1]$ & 0.4 \\
Yaw rate & $r_o = \frac{r}{r_{max}} \in [-1,1]$ & 0.4 \\
Course error & $\tilde{\chi}_o = \frac{\chi_d - \chi}{\chi_{max}} \in [-1,1]$ & $\pi$ \\
Elevation error & $\tilde{\upsilon}_o = \frac{\upsilon_d - \upsilon}{\upsilon_{max}} \in [-1,1]$ & $\pi$ \\
Ocean current velocity, surge & $u_{c,o} = \frac{u_c}{V_{c,max}} \in [-1,1]$ & $1$ \\
Ocean current velocity, sway & $v_{c,o} = \frac{v_c}{V_{c,max}} \in [-1,1]$ & $1$ \\
Ocean current velocity, surge & $w_{c,o} = \frac{w_c}{V_{c,max}} \in [-1,1]$ & $1$
\end{tabular}
\label{tab:pf_e2e_obs}
\end{table}
\endgroup

The neural networks also observe a flattened sonar data image. The raw sonar image is of dimension $(15, 15)$, but to reduce the number of computations needed, dimensionality reduction to $(8,8)$ is performed by minimum pooling. 

\section{Simulation Results}
\subsection{Quantitative Results}
Three values for the trade-off parameter $\lambda_r$ was used during training to obtain three expert level controllers. The quantitative results are obtained by running each training scenario, which are configured randomly in each episode, for $N=100$ episodes. As metrics we use success rate, collision rate and average tracking error over all episodes. Success is defined as the agent reaching the last waypoint within an acceptance radius, i.e. $||p_{t_{final}}-p_{target}||^2 < d_a$ where $d_a$ has been set to $1m$, without colliding. Equivalently, a collision has happened if the distance between the AUV and any obstacle, at any time during an episode, is less than a specified safety radius $d_{safety}=1m$. 

By running controllers trained with different values for $\lambda_r$, one can hypothesize of the outcome of the tests by the incentive the agent has been training with and compare with experimental results. Intuitively, we should see a higher collision rate and lower average tracking error for a controller with high $\lambda_r$, because it has a larger incentive to stay on path. Conversely, the expected outcome of a controller trained with small $\lambda_r$ should yield a higher average tracking error and a lower collision rate. \autoref{tab:quantitative_results} lists the full report from the quantitative tests. 

\begin{table}[H]
    \centering
    \vspace{-0.3in}
    \caption{Test results from sampling $N=100$ random training scenarios.}
    \begin{tabular}{c l c c c c c}
    \toprule
        \textbf{Trade-off} & \textbf{Metric} & \textbf{Intermediate} & \textbf{Proficient} & \textbf{Advanced} & \textbf{Expert} & \textbf{Avg.}\\ \hline
        & \text{Success rate [\%]} & 68 & 66 & 62 & 52 & 62\\
        $\lambda_r=0.9$ & \text{Collision rate [\%]} & 16 & 28 & 34 & 38 & 29 \\
        & \text{Avg. tracking error} & 1.67 & 2.91 & 3.14 & 3.09 & 2.70 \\ \hline
        & \text{Success rate [\%]} & 100 & 100 & 86 & 59 & 86\\
        $\lambda_r=0.5$ & \text{Collision rate [\%]} & 0.00 & 0.00 & 8.00 & 36.0 & 11\\ 
        & \text{Avg. tracking error [m]} & 1.97 & 3.76 & 4.44 & 4.33 & 3.63 \\ \hline
                & \text{Success rate [\%]} & 65 & 68 & 45 & 54 & 54 \\
        $\lambda_r=0.1$ & \text{Collision rate [\%]} & 0 & 0 & 0 & 3 & 0.75 \\ 
        & \text{Avg. tracking error [m]} & 3.98 & 6.15 & 7.91 & 7.33 & 6.34 \\ \hline
    \end{tabular}
    \label{tab:quantitative_results}
\end{table}
\vspace{-0.2in}
The quantitative results can be extrapolated to find general expressions for the success rate, collision rate and average tracking error as functions of $\lambda_r$. The collision rate and the average tracking error are well-described by exponential functions $y=ae^{bx}+c$. It is also seen that a quadratic function $y=ax^2+bx+c$ describes the success rate as a function of the trade-off parameter quite well. This matches our expectations as higher $\lambda_r$ induce more collisions and therefore lowers the success rate. On the other hand, during the episodes where it manages to avoid collisions it always succeeds because the tracking error is very low. Lower $\lambda_r$ configurations naturally has the opposite problem: The low collision rate is due to it being more willing to go off-track, but makes it less likely to reach the end-goal within the acceptance radius. 

\begin{figure}[H]
    \centering
    \includegraphics[width=0.75\linewidth]{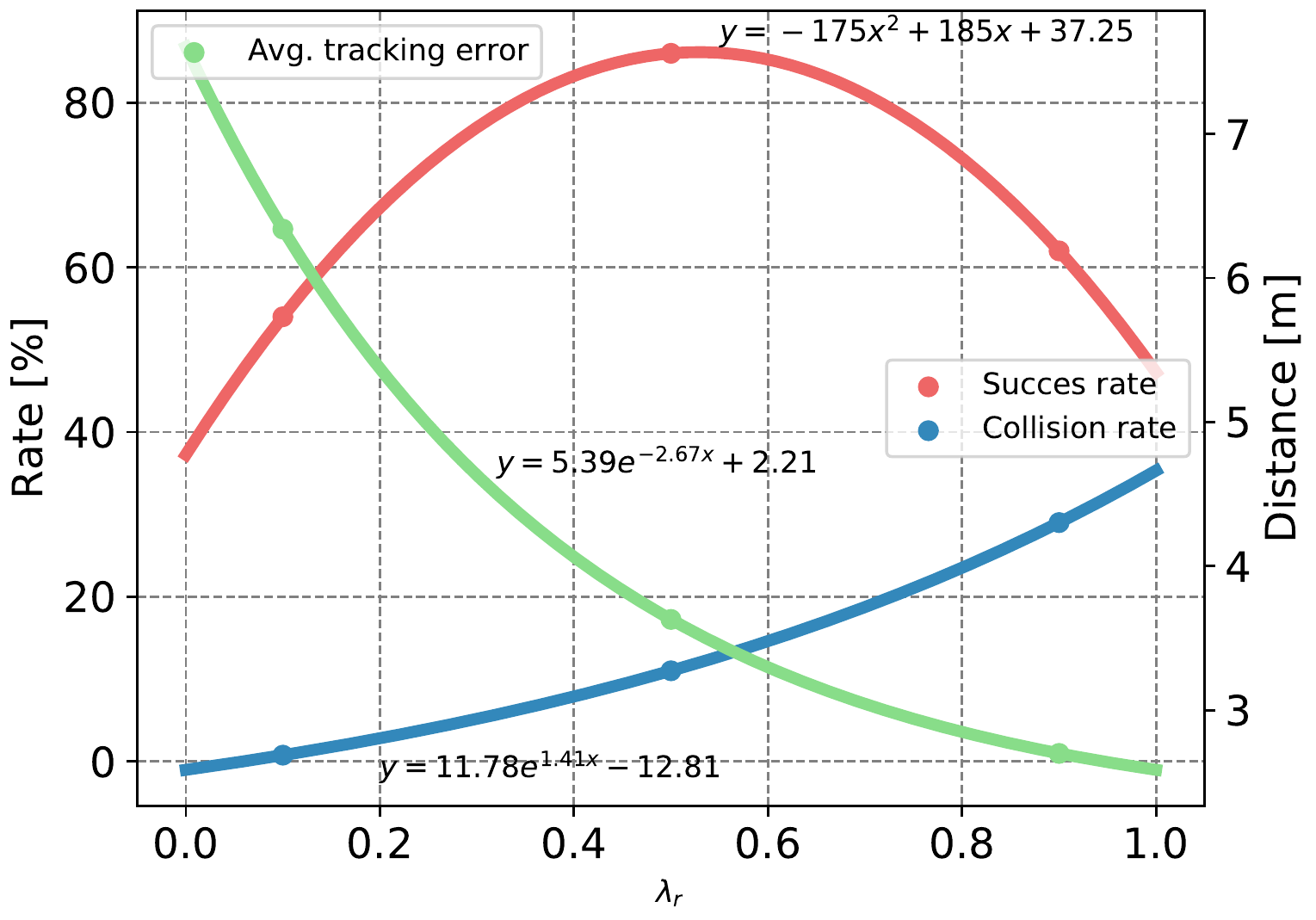}
    \caption{Curve-fitted data from \autoref{tab:quantitative_results}. The average tracking error and the collision fitted to exponential functions, while the success rate is fitted to a quadratic polynomial.}
    \label{fig:my_label}
\end{figure}

\subsection{Qualitative Results}\label{ssec:qual_result}
\begin{figure}[H]
    \centering
    \includegraphics[width=0.9\linewidth]{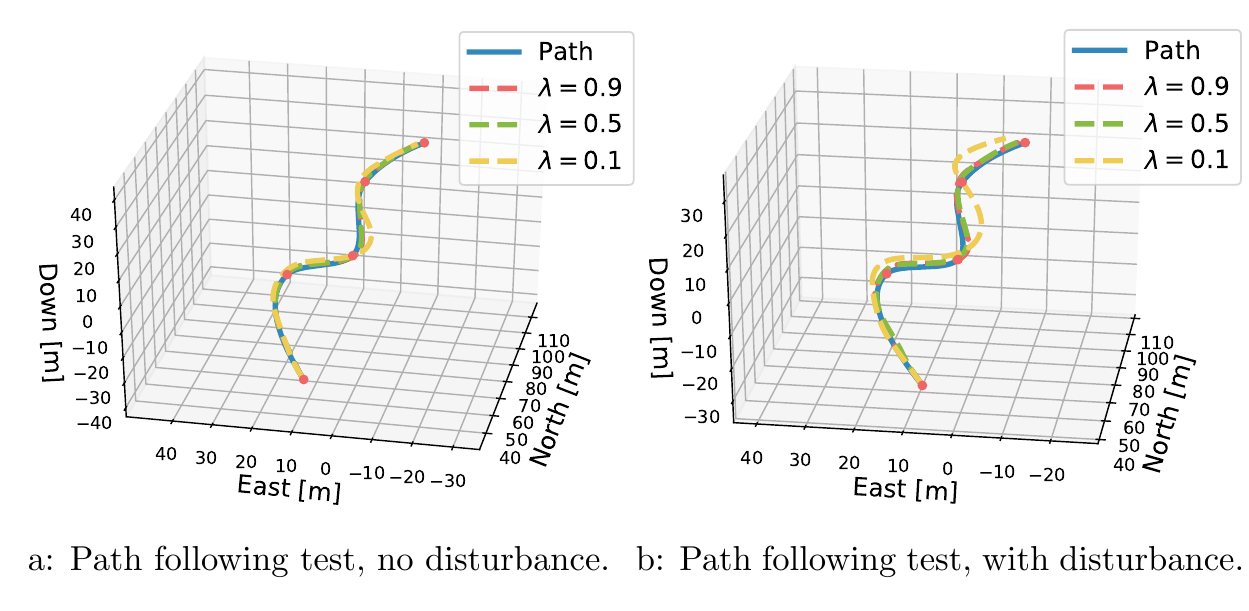}
    \caption{The pure path-following test. As expected, higher $\lambda_r$ are better at path-following. }
    \label{fig:test_pf}
\end{figure}

In the qualitative tests, 4 different scenarios are set-up in order to test different behavioural aspects of the controllers. The first test see the controllers tackle a pure path following test, both with and without the presence of an ocean current. \autoref{fig:test_pf} plots the the results of from simulating one episode. 

The data from this test is summarized in \autoref{tab:pf_results}. When testing, all controllers are run in deterministic mode to ensure that all results are reproducible. For the same reason the current is fixed at a constant intensity and direction. From the test we obtain the same performance observed in the quantitative tests.  The agent tuned with $\lambda_r=0.9$ manages to obtain an average tracking error as low as $0.45m$ in ideal conditions, showcasing impressive tracking on curved 3D paths. Further, it is observed that the tracking errors increases significantly from $0.5$ to $0.1$. This is also reflected in the sensitivity on tracking error due to the presence of the disturbance. Most of the error happens where the path curvature is high. In addition, all cases are successful, except $\lambda_r=0.1$ with current disturbance, which is visibly off-track as it passes the last waypoint.

\begin{table}[H]
    \centering
    \caption{Performance on pure path-following in terms of avg. tracking error}
    \begin{tabular}{c c c c}
    \toprule
        \textbf{Trade-off} & \textbf{Ideal} & \textbf{Perturbed} & \textbf{Disturbance sensitivity}\\ \hline
        $\lambda_r=0.9$ & $0.45m$ & $0.52m$ & $15\%$ \\
        $\lambda_r=0.5$ & $0.54m$ & $0.81m$ & $81\%$ \\
        $\lambda_r=0.1$ & $1.64m$ & $3.95m$ & $141\%$
    \end{tabular}
    \label{tab:pf_results}
\end{table}

\begin{figure}[H]
    \centering
    \includegraphics[width=0.55\linewidth]{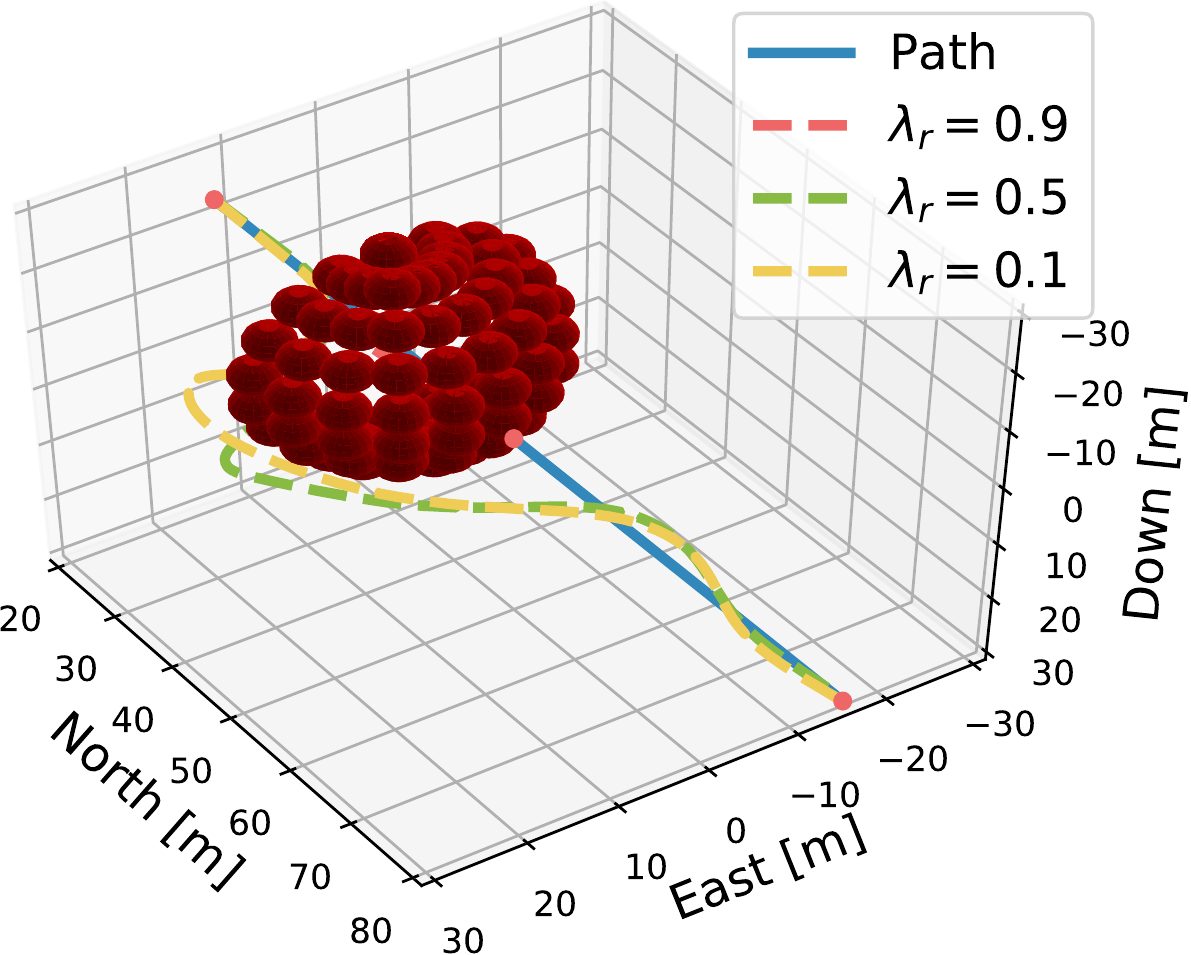}
    \caption{A dead-end test, where the the obstacles are configuration as a half-sphere with a radius of $20m$.}
    \label{fig:deadend}
\end{figure}

Next test involves a dead-end scenario, where obstacles are configured as a half-sphere with radius 20m. This means that the agent will sense the dead-end $5m$ prior to the center, due to the sonar range of $25m$, and must take the appropriate actions to escape it. The test, figured in \autoref{fig:deadend}, indicates that $\lambda_r=0.9$ fails this test and can not escape the dead-end on account of it being too biased to staying on-path. On the other hand, $\lambda_r=0.5, 0.1$ behaves somewhat similarly and manages to escape and reach the goal position. 

\begin{figure}[H]
    \centering
    \includegraphics[width=0.95\linewidth]{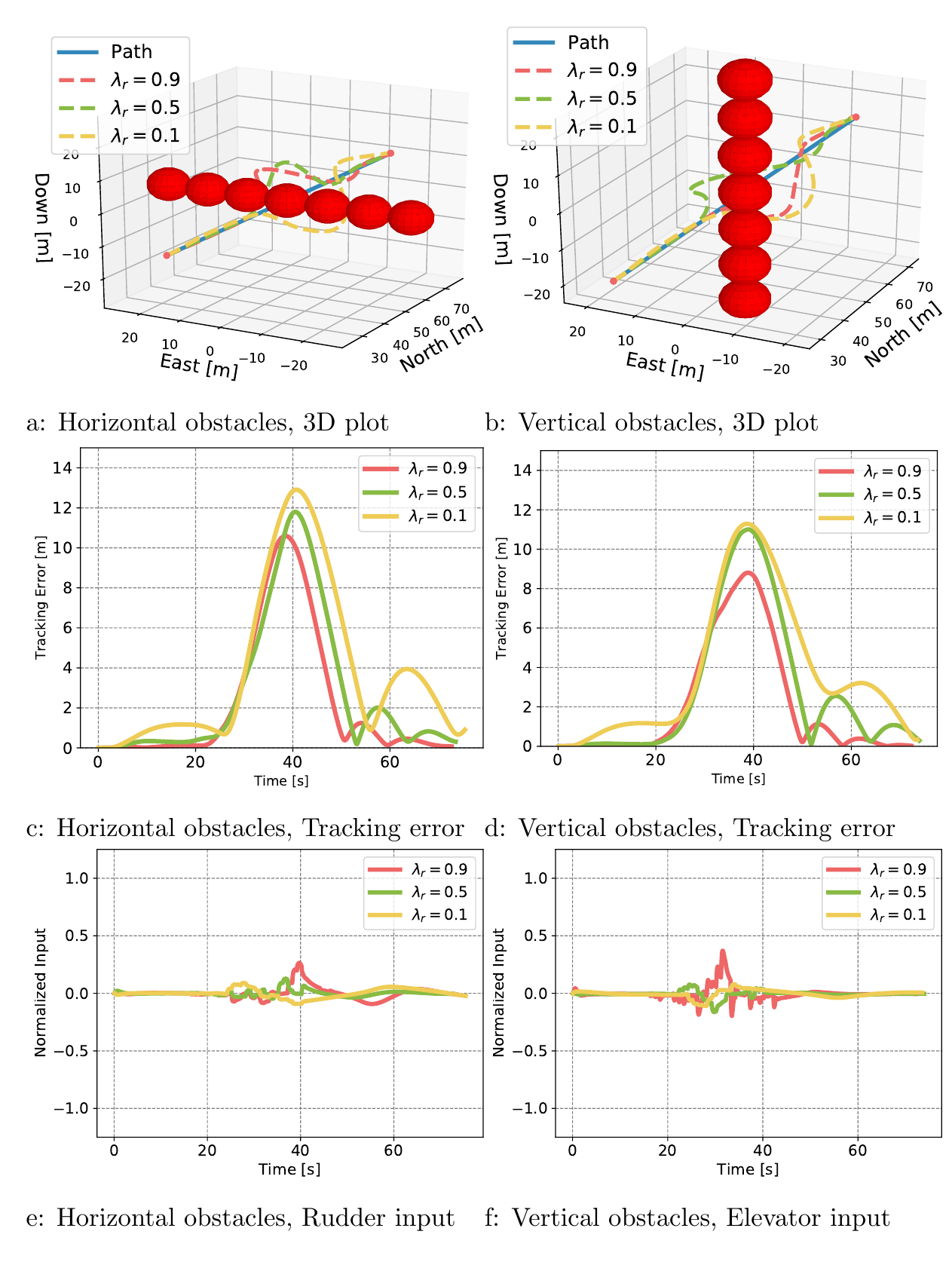}
    \caption{The horizontal and vertical obstacle test. Here, we are interested in seeing if the agent has learned which actuator to use to avoid the obstacles.}
    \label{fig:ver_hor_test}
\end{figure}

In the last test we dissect if the agent learned to operate the actuators effectively according to how obstacles are configured. In the extreme cases, obstacles would be stacked horizontally and veritcally, and optimally no control energy should be spent on taking the AUV towards "the long way around". Instead it should use the actuator to avoid on the lateral side of the stacking direction. As is seen in \autoref{fig:ver_hor_test}, agents behave according to the quantitative results in terms of tracking error. Moreover, all achieves success as they reach the end-goal within the radius of acceptance. It is also seen that the``opposite" control fin are operated very conservatively, as desired. 
\subsection{Analysis}
The results obtained from the test scenarios demonstrates a clear connection to the reward function, as intended. In pure path-following test, the agent biased towards path-following manages to track the path with great precision. On the other hand, regulating the trade-off closer to COLAV, yields agents that are willing to go further off-track to find safe trajectories. This is reflected in the average tracking error and in the collision rate. 

Furthermore, it is seen that the latter controllers seem to react by spending less aggressive control. The controller tuned with $\lambda_r=0.5$ is seen to be effective in avoiding the obstacles and is also not deviating towards the sub-optimal dimension. The expert level agent tuned with $\lambda_r=0.1$ shows great caution and from the quantitative analysis shows $99.25\%$ collision-free samples out of $400$, where collisions occurred at expert level difficulty only. 

A current limitation in the simulated setup, is the assumption that all states, including the ocean current, is available for feedback. We have therefore omitted the \textit{navigation} part of the classical feedback loop for marine crafts. In a full-scale test, state estimation and sensor noise would naturally be part of the feedback-loop, necessitating the need for a navigation module.

\section{Conclusion}
In this research, a deep reinforcement learning agent was trained and deployed to tackle the hybrid objective of 3D path-following and 3D collision avoidance by an autonomous underwater vehicle. Specifically, the state-of-the-art learning algorithm Policy Proximal Optimization was used to train the neural networks. The AUV was operated by commanding three actuator signals in the form of propeller shaft speed and rudder and elevator fin angles. A PI-controller maintained a desired cruise speed, while the DRL agent operated the control fins. The agent took decisions by observing the state variables of the dynamical model, control errors, the disturbances and through sensory inputs from a forward looking sonar. 

A reward system based on quadratic penalization was designed to incentivize the agent to follow the path, but also be willing to deviate if further on-path progress is unsafe. In addition, avoiding excessive roll and use of control actuation was avoided by penalizing such behaviour. As path-following and avoiding collisions are competing objectives, the agent must trade-off one for the other in order to achieve a successful outcome in an episode. Since this trade-off is non-trivial, a regulating parameter $\lambda_r$ was introduced and tuned with three different values to observe behavioural outcome. Furthermore, the three trained controllers were evaluated quantitatively using statistical averages by sampling $N=100$ episodes per difficulty level and measuring the success rate (reaching the last waypoint within an acceptance radius without collision), collision rate and average tracking error. Lastly, the controllers were tested in special-purposed scenarios to investigate the quality of path-following in the special case where no objects are restricting the path, optimal use of actuators in extreme obstacle configurations and in a dead-end test.

From the test data we observed that the trade-off tuning path-following/COLAV bias confirmed the intended relationship from the reward function design. The agent biased towards path-following could follow a track with an average error $<1m$ even in the presence of a perturbing ocean current. Agents biased towards COLAV demonstrated great collision avoidance under ideal conditions, where the best agent demonstrated zero collisions out of $300$ samples. The obtained results indicate that RL could play a part in achieving truly autonomous vehicles capable of human-level decision-making.
\bibliographystyle{abbrvnat}
\bibliography{references}  
\end{document}